\documentclass[letterpaper]{article} 
\usepackage{aaai2027}  
\usepackage[hyphens]{url}  
\usepackage{graphicx} 
\usepackage{natbib}  
\usepackage{caption} 
\usepackage{algorithm}
\usepackage{algorithmic}
\usepackage{colortbl}
\usepackage{xcolor}
\usepackage{makecell}
\usepackage{amsmath}

\usepackage{newfloat}
\usepackage{listings}
\DeclareCaptionStyle{ruled}{labelfont=normalfont,labelsep=colon,strut=off} 
\floatstyle{ruled}
\newfloat{listing}{tb}{lst}{}
\floatname{listing}{Listing}

\usepackage{booktabs}

\title{CONFER: Conflict-Aware Evidence Negotiation for Regime-Calibrated Weak Supervision in Multimodal Emotion Recognition}
\title{CONFER: Conflict-Aware Evidence Negotiation for Regime-Calibrated Weak Supervision in Multimodal Emotion Recognition}

\author{
    Bojing Hou,
    Ruohao Li,
    Yitong Zhu,
    Luwen Yu,
    Yuyang Wang\corresponding
}

\affiliations{
    The Hong Kong University of Science and Technology (Guangzhou)\\
    bhou870@connect.hkust-gz.edu.cn,
    rli777@connect.hkust-gz.edu.cn,\\
    yzhu162@connect.hkust-gz.edu.cn,
    luwenyu@hkust-gz.edu.cn,
    yuyangwang@hkust-gz.edu.cn
}

\begin{document}

\maketitle

\begin{abstract}

Multimodal emotion recognition often treats self-reported labels as reliable supervision while overlooking self-report unreliability and cross-modal conflict. We propose \textbf{CONFER}, a graph-based conflict-aware evidence negotiation framework for weakly supervised multimodal emotion recognition. CONFER represents each modality expert as a node with a predictive belief, boundary-based uncertainty, and runtime reliability estimated from historical out-of-fold performance and current-sample uncertainty. Uncertainty-aware compatibility and reliability-directed asymmetric edge weights govern iterative message-passing negotiation, followed by peer-supported prediction readout. Conflict reduction, residual disagreement, and mean modality uncertainty further characterize three regimes---Consensus, Dissent, and Ambiguity---for sample-specific weak-label calibration. We evaluate CONFER on AMIGOS, MAHNOB-HCI, and DEAP under subject-dependent 10-fold and strict leave-one-subject-out (LOSO) protocols. CONFER achieves competitive performance, reaching \textbf{0.873} accuracy on AMIGOS-V and \textbf{0.854} accuracy on MAHNOB-V under strict LOSO evaluation. Further analyses show larger negotiation gains on high-conflict samples and improved robustness to weak-label corruption, indicating that cross-modal conflict provides useful information for both directional modality coordination and supervision-reliability estimation.
\end{abstract}

\section{Introduction}

Multimodal emotion recognition (MER) aims to infer latent emotional states from heterogeneous neural, physiological, acoustic, and behavioral signals. Although these modalities provide complementary evidence, most existing methods only treat self-reported valence and arousal scores as reliable supervision. In practice, self-reports are influenced by individual rating habits, cognitive interpretation, memory bias, and experimental context~\cite{mauss2009measures,barrett2017constructed}, and are therefore better regarded as weak observations of latent emotion.

This challenge is further complicated by \textbf{cross-modal conflict}, which we define as 
disagreement among modality-specific predictions. Such conflict may arise from heterogeneous sources of unreliability across modalities (e.g., sensing noise, modality-specific limitations, or insufficient evidence)~\cite{bezirganyan2025discounted,lu2025navigating}.
However, cross-modal conflict can also reveal complementary evidence and guide selective knowledge exchange between modalities.
Yet discrepancies between fused predictions and self-reported labels alone cannot determine whether the inconsistency arises from unreliable modality predictions or weak labels. MER must therefore jointly model modality reliability, uncertainty and cross-modal conflict to determine when agreement or conflict between modalities is informative and how weak-label supervision should be calibrated.



Existing MER methods address modality heterogeneity through attention, gating, cross-modal interaction, 
shared and modality-specific representation learning, and expert routing, 
but forcibly unify conflicts between modalities through fusion weights,
without explicitly determining whether one modality should influence another for the current sample~\cite{han2024fusemoe,xin2025i2moe}. Meanwhile, weak-label methods estimate supervision reliability from confidence, loss, or global statistics 
while overlooking pairwise agreement and conflict among modality-specific predictions.
In existing frameworks, modality reliability estimation, cross-modal conflict modeling, and weak-label calibration remain disconnected. 
Consequently, information passing between modalities cannot be efficiently orchestrated or robustly learned.

To address this gap, we propose \textbf{CONFER} (\textbf{CONF}lict-Aware \textbf{E}vidence Negotiation for \textbf{R}egime-Calibrated Weak Supervision). Each modality expert produces a predictive belief and a boundary-based uncertainty score, while its historical out-of-fold performance is combined with current uncertainty to estimate runtime reliability. Together, these quantities constitute modality evidence. Inspired by multi-agent negotiation, CONFER represents modality experts as nodes in a dynamic graph, where pairwise compatibility and reliability-aware directional edge weights govern iterative message passing. A peer-supported readout produces the final multimodal prediction, while residual conflict and uncertainty characterize three regimes---\emph{Consensus}, \emph{Dissent}, and \emph{Ambiguity}---to calibrate each sample's contribution to weak-label training. Thus, CONFER connects modality reliability, cross-modal negotiation, and weak-label calibration within a unified framework.


We evaluate CONFER on AMIGOS, MAHNOB-HCI, and DEAP under subject-dependent 10-fold and strict LOSO protocols. CONFER achieves competitive performance, including 0.873 accuracy/0.879 F1 on AMIGOS-V and 0.854 accuracy on MAHNOB-V under LOSO. 
Further analyses show larger gains on high-conflict samples, improved robustness under synthetic label corruption, and supervision-reliability estimates that provide information beyond predictive confidence.

The main contributions are summarized as follows:

\begin{itemize}






\item We propose \textbf{CONFER}, a unified graph-based framework that negotiates cross-modal conflict according to sample-specific modality reliability and calibrates learning from self-reported weak labels.




\item We develop a dynamic graph mechanism that converts uncertainty-aware compatibility into reliability-directed asymmetric message passing, allowing cross-modal conflict to guide both expert coordination and supervision-reliability estimation.

\item Systematic experiments on three MER benchmarks demonstrate that CONFER achieves competitive cross-subject performance and improved robustness to weak-label corruption.

\end{itemize}

\section{Related Work}

Multimodal emotion recognition integrates visual, acoustic, neural, and physiological signals~\cite{baltrusaitis2019multimodal,mittal2020m3er,rayatdoost2020gated,zhang2021regularized}. Representative fusion methods include TFN, MulT, MISA, Self-MM, and MMIM~\cite{zadeh2017tfn,tsai2019mult,hazarika2020misa,yu2021selfmm,han2021mmim}, while recent expert-routing approaches adapt modality contributions to individual inputs~\cite{han2024fusemoe,fang2025emoe,xin2025i2moe,gao2024euar}. However, these methods mainly optimize feature aggregation or modality weighting without explicitly relating cross-modal conflict to supervision reliability.

Modality reliability varies with noise, missing information, and individual differences~\cite{liu2024missing,zeng2022mitigating}, and has been modeled through attention, gating, confidence, uncertainty estimation, and uncertainty-aware fusion~\cite{guo2017calibration,kendall2017uncertainties,sensoy2018evidential,han2021tmc,han2023dynamic}. Meanwhile, self-reported emotion labels are affected by subjective perception and experimental conditions~\cite{mauss2009measures,barrett2017constructed}. Noisy-label methods employ robust losses, reweighting, co-training, sample selection, and label correction~\cite{song2023noisylabels,zhang2018gce,wang2019sce,han2018coteaching,li2020dividemix,shu2023cmwnet,wang2024noisegpt}, typically estimating supervision reliability from confidence, loss, training dynamics, or representation neighborhoods~\cite{liu2020elr,xia2022uncertainty,li2022selcl,huang2023twin,kim2024confrag}. These directions rarely exploit pairwise cross-modal relationships to jointly coordinate modality interaction and calibrate weak-label supervision, which is the focus of CONFER.

\begin{figure*}[t]
    \centering
    \includegraphics[width=\textwidth]{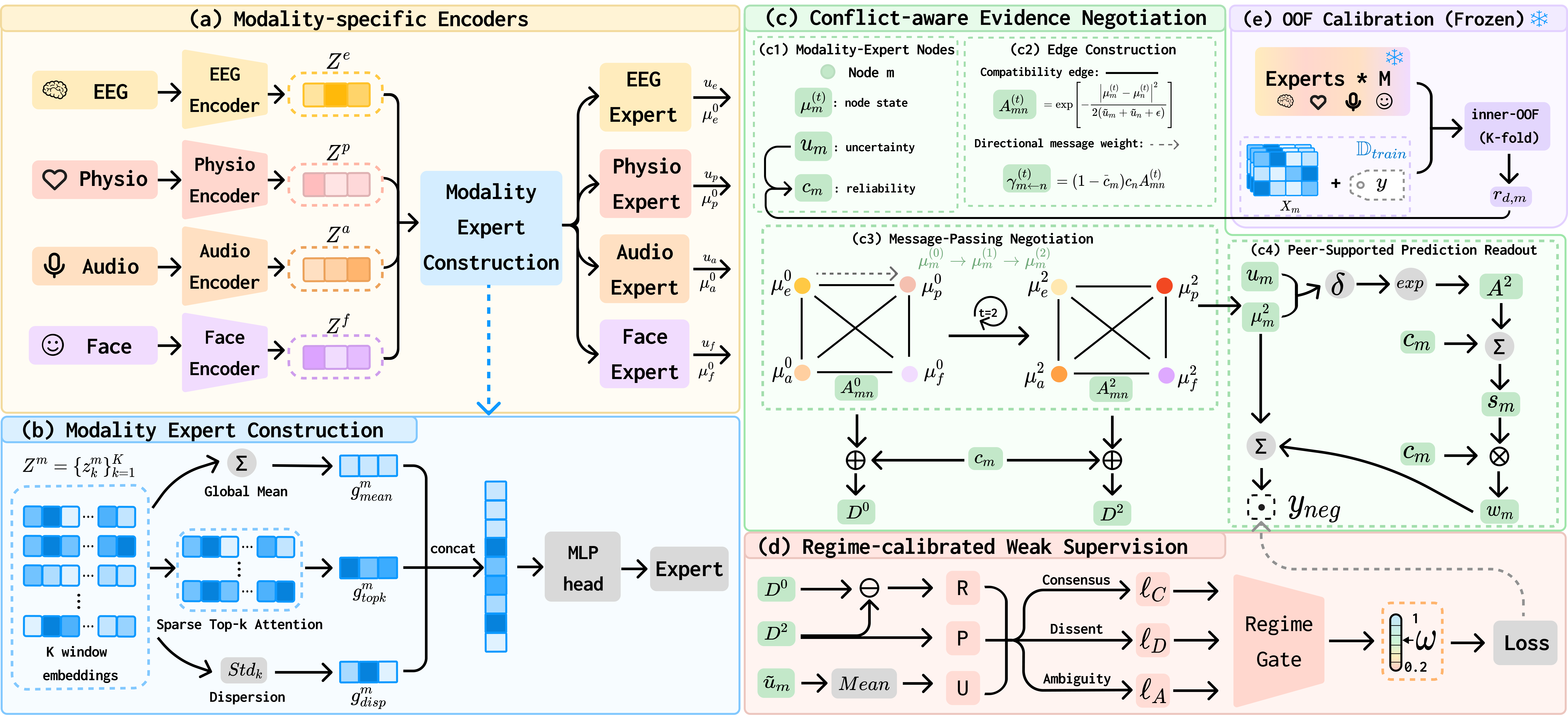}
\caption{
Overview of the CONFER framework.
(a) Modality-specific encoders extract window-level representations.
(b) Modality experts produce trial-level beliefs and boundary-based uncertainty scores.
(c) Dynamic compatibility and directional edges govern message-passing negotiation and peer-supported prediction readout.
(d) Conflict and uncertainty guide regime-calibrated weak supervision.
(e) Frozen inner-OOF statistics provide historical reliability priors.
}
    \label{fig:CONFER_framework}
\end{figure*}

 \begin{table}[t]
  \centering
  \caption{
  Modality-wise prediction error and uncertainty statistics on AMIGOS-V under strict-LOSO evaluation.
  }
  \label{tab:modality_reliability}
  \small
  \setlength{\tabcolsep}{4pt}
  \begin{tabular}{lcccc}
  \toprule
  \textbf{Modality}
  & \textbf{Mean Error} $\downarrow$
  & \textbf{Error Std}
  & \textbf{Mean $\widetilde{u}_m$} $\downarrow$
  & \textbf{Median $\widetilde{u}_m$} \\
  \midrule
  Audio  & 0.225 & 0.331 & 1.22 & 1.00 \\
  Face   & 0.263 & 0.499 & 1.55 & 1.02 \\
  EEG    & 0.350 & 0.497 & 2.10 & 0.98 \\
  Physio & 0.375 & 0.499 & 2.00 & 1.01 \\
  \bottomrule
  \end{tabular}
  \end{table}

\section{Method}

\subsection{Overview and Problem Formulation}

Multimodal emotion recognition (MER) aims to infer latent emotional states from heterogeneous neural, physiological, acoustic, and behavioral observations. Given a trial $i$, its multimodal input is represented as
\begin{equation}
\begin{array}{rcl}
\mathcal{X}_i
& = &
\left\{
X_i^m
\right\}_{m\in\mathcal{M}},
\\[4pt]
X_i^m
& = &
\left\{
x_{i,k}^m
\right\}_{k=1}^{K_m}.
\end{array}
\label{eq:problem_formulation}
\end{equation}

Here, \(\mathcal{M}\) is the modality set, \(x_{i,k}^m\) is the \(k\)-th window of modality \(m\), and \(K_m\) is the number of windows. Each trial has a self-reported weak label \(y_i^w\). We formulate MER as jointly estimating sample-dependent modality reliability, negotiating cross-modal conflict, and calibrating weak-label supervision.

As shown in Fig.~\ref{fig:CONFER_framework}, CONFER comprises modality-expert construction, dynamic-graph negotiation, and regime-calibrated weak supervision. Historical reliability and calibration statistics are estimated from frozen inner-OOF predictions using only outer-training subjects. The trial index \(i\) is omitted when unambiguous.




\subsection{Modality Expert Construction}

A fixed modality weight assumes that modality quality remains stable across subjects and samples. However, Table~\ref{tab:modality_reliability} shows clear differences in prediction error and uncertainty across modalities under strict leave-one-subject-out evaluation. We therefore model each modality as an independent expert whose reliability depends on both historical performance and the current sample.



As illustrated in Fig.~\ref{fig:CONFER_framework}(a), the modality-specific encoder maps the temporal windows of modality \(m\) to an embedding sequence \(Z^m=\{z_k^m\}_{k=1}^{K_m}\). The modality expert constructor in Fig.~\ref{fig:CONFER_framework}(b) then applies global mean pooling to capture persistent trial-level responses, sparse top-\(k\) attention to preserve informative temporal events, and temporal dispersion to characterize variations hidden by average pooling. Their outputs, \(g_{\mathrm{mean}}^m\), \(g_{\mathrm{topk}}^m\), and \(g_{\mathrm{disp}}^m\), are concatenated into the trial-level representation \(g_m\).

A belief head produces the initial predictive belief, from which the boundary-based uncertainty score is computed:
\begin{equation}
\begin{array}{rcl}
\mu_m^{(0)}
& = &
\sigma
\left(
h_\mu^m(g_m)
\right),
\\[6pt]
u_m
& = &
\mu_m^{(0)}(1-\mu_m^{(0)})
+\epsilon .
\end{array}
\label{eq:expert_state}
\end{equation}

Here, \(h_\mu^m\) is the belief head, \(\sigma(\cdot)\) is the sigmoid function, and \(\epsilon>0\) is a small constant for numerical stability. The initial predictive belief is denoted by \(\mu_m^{(0)}\), while \(u_m\) is a boundary-based uncertainty score whose value increases as the prediction approaches the decision boundary.

\subsection{Conflict-Aware Evidence Negotiation}


CONFER represents modality experts as nodes in a dynamic negotiation graph. At negotiation round \(t\), the predictive belief \(\mu_m^{(t)}\) serves as the state of node \(m\), while its boundary-based uncertainty \(u_m\) and runtime reliability \(c_m\) characterize the reliability of that state. Pairwise compatibility \(A_{mn}^{(t)}\) defines the graph edges, and receiver-specific message weights \(\gamma_{m\leftarrow n}^{(t)}\) govern directional information exchange and node updates. The negotiated node states are subsequently read out to produce the multimodal prediction, as illustrated in Fig.~\ref{fig:CONFER_framework}(c).
Reliability estimation alone cannot resolve sample-level disagreement, because predictions from different experts should not be directly combined without considering whether their disagreement is consistent with their uncertainty.

\subsubsection{Modality-Expert Node State and Reliability Characterization}

As summarized in Fig.~\ref{fig:CONFER_framework} (c1), each modality-expert node is characterized by its predictive state, boundary-based uncertainty, and runtime reliability.
To capture both historical modality quality and sample-specific uncertainty, runtime reliability is estimated by combining the OOF reliability prior with current uncertainty.
Specifically, \(r_{d,m}\) is estimated as the inner-OOF balanced accuracy of modality \(m\) on the outer-training subjects, where \(d\) denotes the dataset--affective-dimension setting. The median boundary-based uncertainty obtained from the same OOF predictions is recorded as \(u_{m,\mathrm{med}}\). At runtime, uncertainty is normalized as \(\widetilde{u}_m=u_m/u_{m,\mathrm{med}}\) and combined with the historical prior:

\begin{equation}
c_m
=
r_{d,m}
\exp
\left(
-\widetilde{u}_m
\right).
\label{eq:runtime_reliability}
\end{equation}


Here, \(c_m\) is the runtime reliability of the expert associated with modality \(m\). It controls the expert's outgoing contribution, while its max-normalized form \(\overline{c}_m\), obtained by dividing \(c_m\) by the largest runtime reliability within the current trial, controls the acceptance of incoming peer beliefs. All OOF statistics are estimated only from the outer-training subjects and fixed before joint optimization, preventing held-out-subject leakage. 
By construction, the median normalized uncertainty over the OOF calibration samples is one for each modality.
Together, \(\mu_m^{(0)}\), \(u_m\), and \(c_m\) constitute the modality evidence used in subsequent negotiation.


\subsubsection{Pairwise Compatibility and Directional Edge Weighting}

as shown in component (c2) of Fig.~\ref{fig:CONFER_framework}, at negotiation round \(t\), a compatibility edge is constructed between each pair of modality-expert nodes \(m,n\in\mathcal{M}\) from their uncertainty-adjusted prediction disagreement:

\begin{equation}
A_{mn}^{(t)}
=
\exp
\left[
-
\frac{
\left|
\mu_m^{(t)}
-
\mu_n^{(t)}
\right|^2
}{
2
\left(
\widetilde{u}_m
+
\widetilde{u}_n
+
\epsilon
\right)
}
\right].
\label{eq:dynamic_compatibility}
\end{equation}

Because \(A_{mn}^{(t)}=A_{nm}^{(t)}\), it represents symmetric pairwise compatibility rather than the direction of information exchange. Self-edges are excluded by setting \(A_{mm}^{(t)}=0\). Compatibility decreases when two confident experts disagree, but remains relatively high when their disagreement can be explained by uncertainty.


The symmetric compatibility scores are converted into receiver-specific directional edge weights:
\begin{equation}
\gamma_{m\leftarrow n}^{(t)}
=
\left(
1-\overline{c}_m
\right)
c_n
A_{mn}^{(t)}.
\label{eq:directional_message}
\end{equation}
Although compatibility is symmetric, the resulting edge weights are generally asymmetric because the sender and receiver reliabilities play different roles.

\subsubsection{Iterative Message-Passing Negotiation}

As shown in component (c3) of Fig.~\ref{fig:CONFER_framework},
for receiver node \(m\), \(\overline{\mu}_m^{(t)}\) is the \(\gamma_{m\leftarrow n}^{(t)}\)-weighted average of the incoming sender states, and \(Z_m^{(t)}\) is their total weight. We set \(\lambda_m^{(t)}=\eta Z_m^{(t)}/(1+Z_m^{(t)})\), where \(\eta\in(0,1]\) bounds the concession strength. The node state is then updated as:
\begin{equation}
\begin{array}{rcl}
\mu_m^{(t+1)}
& = &
\left(
1-\lambda_m^{(t)}
\right)
\mu_m^{(t)}
\\[4pt]
& & {}+
\lambda_m^{(t)}
\overline{\mu}_m^{(t)}.
\end{array}
\label{eq:node_state_update}
\end{equation}
When \(Z_m^{(t)}=0\), the peer aggregation is omitted and \(\lambda_m^{(t)}=0\), so node \(m\) retains its current state.
Reliable senders contribute more, whereas less reliable receivers accept more information from compatible peers. We perform two negotiation rounds, \(\mu_m^{(0)}\rightarrow\mu_m^{(1)}\rightarrow\mu_m^{(2)}\), recomputing compatibility and directional edge weights after each update. The final graph state \(\{\mu_m^{(2)},u_m,c_m,A_{mn}^{(2)}\}\) is used for prediction readout and conflict-regime characterization.

\subsubsection{Peer-Supported Multimodal Prediction Readout}

As shown in of Fig.~\ref{fig:CONFER_framework} (c4), the negotiated graph state is read out to produce the final multimodal prediction.
Individual reliability cannot solely determine whether an expert is supported by the remaining modalities. We therefore define $s_m$ as the final compatibility between the expert of modality \(m\) and the experts of the other modalities, averaged using the corresponding peer reliabilities \(c_n\) as weights. Its unnormalized readout weight is
\begin{equation}
\widetilde{w}_m
=
c_m
\left[
\beta
+
\left(
1-\beta
\right)
s_m
\right].
\label{eq:readout_weight}
\end{equation}

The unnormalized readout weights \(\widetilde{w}_m\) are normalized across modalities to obtain \(w_m\),
and the final negotiated prediction is
\begin{equation}
y_{\mathrm{neg}}
=
\sum_{m\in\mathcal{M}}
w_m
\mu_m^2.
\label{eq:negotiated_prediction}
\end{equation}

The peer-support term assigns greater readout weights to nodes that are compatible with other reliable experts,
while $\beta\in[0,1]$ preserves individual reliability and prevents a reliable expert from being discarded solely because it disagrees with the majority.


\subsection{Regime-calibrated Weak Supervision}

The conflict-regime inference and regime-based sample-weighting process are illustrated in Fig.~\ref{fig:CONFER_framework} (d).

\subsubsection{Conflict Regime Inference}

The negotiation outcome is characterized by conflict reduction, residual disagreement, and overall uncertainty. 
At round \(t\), \(D^{(t)}\) is the \(c_mc_n\)-weighted mean of pairwise incompatibility \(1-A_{mn}^{(t)}\) over unordered node pairs. The regime factors are then defined from \(D^{(0)}\), \(D^{(2)}\), and modality uncertainty:
\begin{equation}
\begin{array}{rcl}
R & = & \max\left(D^{(0)}-D^{(2)},0\right),
\\[6pt]
P
& = &
D^{(2)},
\\[6pt]
U
& = &
\displaystyle
\frac{1}{|\mathcal{M}|}
\sum_{m\in\mathcal{M}}
\widetilde{u}_m.
\end{array}
\label{eq:regime_factors}
\end{equation}

Here, \(R\), \(P\), and \(U\) represent resolved conflict, persistent conflict, and mean modality uncertainty, respectively.


The regime factors are mapped to the Consensus, Dissent, and Ambiguity regimes through structured logits:
\begin{equation}
\begin{array}{rcl}
\ell_C
& = &
a_C
\left(
1-P
\right)
+
b_C R
-
c_C U,
\\[6pt]
\ell_D
& = &
a_D P
-
b_D U,
\\[6pt]
\ell_A
& = &
a_A U.
\end{array}
\label{eq:regime_logits}
\end{equation}

The coefficients \(a_C\), \(b_C\), \(c_C\), \(a_D\), \(b_D\), and \(a_A\) are learned during OOF calibration and constrained to be positive through a softplus parameterization.
A softmax produces the regime probabilities $\pi_C$, $\pi_D$, and $\pi_A$, corresponding to Consensus, Dissent, and Ambiguity, respectively.

\subsubsection{Regime-Based Sample Weighting}

The inferred regime probabilities do not modify the self-reported label, but determine the contribution of each trial:
\begin{equation}
\omega_i
=
\pi_{i,C}
+
\alpha_D \pi_{i,D}
+
\alpha_A \pi_{i,A}.
\label{eq:sample_weight}
\end{equation}

We set \(\alpha_D=0.5\) and \(\alpha_A=0.2\), satisfying \(1>\alpha_D>\alpha_A>0\), to reflect the supervision-reliability ordering Consensus \(>\) Dissent \(>\) Ambiguity.
This design keeps $\omega_i$ non-zero for ambiguity samples while reducing their supervision contribution.

The weak-label loss $\mathcal{L}_{\mathrm{weak}}$ is the $\omega_i$-weighted mean BCE between $y_{i,\mathrm{neg}}$ and $y_i^w$. 
When computing this loss, \(\operatorname{stopgrad}(\omega_i)\) is used to prevent the model from reducing the objective by artificially increasing uncertainty or assigning difficult samples to Ambiguity.

\begin{table*}[t]
\centering
\caption{
Subject-independent LOSO results (mean$\pm$std across held-out subjects).
}
\label{tab:loso_main}

\begin{tabular}{llcccccc}
\toprule
\textbf{Method} 
& \textbf{Metric}
& \textbf{AMIGOS-V}
& \textbf{AMIGOS-A}
& \textbf{MAHNOB-V}
& \textbf{MAHNOB-A}
& \textbf{DEAP-V}
& \textbf{DEAP-A}
\\
\midrule

{Late Fusion}
& Acc
&0.858$\pm$0.111
&0.612$\pm$0.213
&0.741$\pm$0.127
&0.671$\pm$0.164
&0.505$\pm$0.107
&\textbf{0.577$\pm$0.147}
\\
\rowcolor{gray!10}
& F1
&0.866$\pm$0.101
&0.653$\pm$0.258
&0.705$\pm$0.197
&0.597$\pm$0.251
&0.562$\pm$0.240
&0.661$\pm$0.229
\\

{TFN}
& Acc
&0.846$\pm$0.113
&0.607$\pm$0.192
&0.778$\pm$0.103
&0.635$\pm$0.174
&0.516$\pm$0.109
&0.527$\pm$0.137
\\
\rowcolor{gray!10}
& F1
&0.854$\pm$0.094
&0.617$\pm$0.246
&0.774$\pm$0.117
&0.570$\pm$0.291
&0.513$\pm$0.251
&0.604$\pm$0.220
\\

{Attention-MIL}
& Acc
&0.842$\pm$0.120
&0.626$\pm$0.197
&0.739$\pm$0.091
&0.622$\pm$0.158
&0.531$\pm$0.100
&0.563$\pm$0.161
\\
\rowcolor{gray!10}
& F1
&0.850$\pm$0.107
&0.648$\pm$0.254
&0.736$\pm$0.135
&0.542$\pm$0.279
&0.569$\pm$0.216
&0.648$\pm$0.233
\\

{MISA}
& Acc
&0.829$\pm$0.144
&0.619$\pm$0.190
&0.698$\pm$0.106
&0.641$\pm$0.157
&0.520$\pm$0.088
&0.536$\pm$0.157
\\
\rowcolor{gray!10}
& F1
&0.835$\pm$0.131
&0.641$\pm$0.245
&0.675$\pm$0.127
&0.582$\pm$0.271
&0.485$\pm$0.284
&0.627$\pm$0.242
\\

{FuseMoE}
& Acc
&0.807$\pm$0.155
&0.617$\pm$0.193
&0.749$\pm$0.113
&0.663$\pm$0.136
&0.522$\pm$0.097
&0.563$\pm$0.150
\\
\rowcolor{gray!10}
& F1
&0.816$\pm$0.150
&0.628$\pm$0.248
&0.743$\pm$0.151
&0.658$\pm$0.211
&0.459$\pm$0.281
&0.665$\pm$0.195
\\

{EMoE}
& Acc
&0.856$\pm$0.116
&0.576$\pm$0.194
&0.728$\pm$0.124
&0.631$\pm$0.126
&0.518$\pm$0.109
&0.541$\pm$0.153
\\
\rowcolor{gray!10}
& F1
&0.862$\pm$0.115
&0.594$\pm$0.254
&0.729$\pm$0.141
&0.563$\pm$0.252
&0.469$\pm$0.302
&0.648$\pm$0.201
\\

{I2MoE}
& Acc
&0.839$\pm$0.130
&0.602$\pm$0.197
&0.751$\pm$0.096
&0.670$\pm$0.161
&0.551$\pm$0.114
&0.550$\pm$0.165
\\
\rowcolor{gray!10}
& F1
&0.852$\pm$0.115
&0.611$\pm$0.246
&0.741$\pm$0.110
&0.664$\pm$0.225
&0.566$\pm$0.231
&\textbf{0.671$\pm$0.186}
\\

{\textbf{CONFER (Ours)}}
& Acc
&\textbf{0.873$\pm$0.122}
&\textbf{0.632$\pm$0.187}
&\textbf{0.854$\pm$0.086}
&\textbf{0.687$\pm$0.153}
&\textbf{0.559$\pm$0.104}
&0.568$\pm$0.133
\\
\rowcolor{gray!10}
& F1
&\textbf{0.879$\pm$0.114}
&\textbf{0.657$\pm$0.226}
&\textbf{0.783$\pm$0.138}
&\textbf{0.679$\pm$0.201}
&\textbf{0.578$\pm$0.270}
&0.650$\pm$0.252
\\

\bottomrule
\end{tabular}

\end{table*}

\subsection{Training and Optimization}

CONFER is trained in three phases. In \textbf{Phase A}, modality-specific encoders and belief heads are trained to produce initial predictive beliefs, from which boundary-based uncertainty scores are deterministically computed.

In \textbf{Phase B}, as illustrated in Fig.~\ref{fig:CONFER_framework} (e), 
the Phase-A experts are retrained on the training portion of each inner split and evaluated on its held-out subjects to obtain OOF predictions.
These predictions are used to estimate \(r_{d,m}\) and \(u_{m,\mathrm{med}}\). After message-passing negotiation, the resulting graph states provide the OOF multimodal prediction \(\hat{y}_i^{\mathrm{OOF}}\) through prediction readout and the regime factors used to compute \(\omega_i\).
The calibration target is defined as
\begin{equation}
\begin{aligned}
t_i
&=
1-\operatorname{Norm}
\left(
\operatorname{BCE}
(\hat{y}^{\mathrm{OOF}}_i,y_i^w)
\right),\\
\tilde{t}_i
&=
\alpha_A+(1-\alpha_A)t_i.
\end{aligned}
\end{equation}
Here, $t_i$ measures OOF prediction--label consistency rather than label correctness, and $\operatorname{Norm}(\cdot)$ denotes min--max normalization over the OOF calibration samples within each outer-training split. The regime coefficients are optimized by minimizing the mean squared error between $\omega_i$ and $\tilde{t}_i$. All calibrated coefficients and OOF statistics are then fixed.

In \textbf{Phase C}, the encoders, belief heads, negotiation module, and prediction-readout module are jointly optimized using
\begin{equation}
\mathcal{L}
=
\mathcal{L}_{\mathrm{weak}}
+
\lambda_{\mathrm{expert}}
\mathcal{L}_{\mathrm{expert}}.
\label{eq:training_objective}
\end{equation}
Here, \(\mathcal{L}_{\mathrm{expert}}\) is the mean BCE over the pre-negotiation modality beliefs, \(\mathcal{L}_{\mathrm{weak}}\) supervises the negotiated prediction using \(\operatorname{stopgrad}(\omega_i)\), and \(\lambda_{\mathrm{expert}}\geq0\) balances the two objectives.

\begin{table}[t]
    \centering
    \caption{Dataset statistics after multimodal alignment.}
    \label{tab:datasets}
    \small
    \setlength{\tabcolsep}{3.5pt}
    \begin{tabular}{lccc}
        \toprule
        \textbf{Dataset}
        & \textbf{Subjects}
        & \textbf{Modalities}
        & \textbf{Task} \\
        \midrule
        AMIGOS
        & 40
        & Face, Audio, EEG, Physio.
        & V / A \\
        
        DEAP
        & 22
        & Face, EEG, Physio.
        & V / A \\
        
        MAHNOB-HCI
        & 28
        & Face, Audio, EEG, Physio.
        & V / A \\
        \bottomrule
    \end{tabular}
\end{table}

\section{Experiments and Results}

\subsection{Experimental Setup}

\subsubsection{Datasets}


We evaluate CONFER on AMIGOS~\cite{miranda2021amigos}, MAHNOB-HCI~\cite{soleymani2012mahnob}, and DEAP~\cite{koelstra2012deap}, which provide self-reported valence/arousal ratings on a 1--9 Self-Assessment Manikin scale~\cite{russell1980circumplex,bradley1994sam}. Table~\ref{tab:datasets} reports the subjects retained after multimodal alignment.

  \subsubsection{Evaluation Protocol}


We use subject-independent LOSO as the primary protocol and additionally report subject-dependent trial-wise 10-fold CV~\cite{kapoor2023leakage}. Labels are binarized using training-fold medians, and accuracy and F1-Score are reported. For each test fold or held-out subject, predictions are averaged over three seeds; standard deviations are then computed across folds or subjects.

  \subsubsection{Implementation Details}

Feature and temporal modalities use gated residual MLPs and multi-scale 1D-CNNs, respectively, with \(d=64\) and 48 windows per trial. We set \(T=2\), \(\eta=0.5\), \(\beta=0.4\), \(\alpha_D=0.5\), \(\alpha_A=0.2\), and \(\lambda_{\mathrm{expert}}=0.15\). AdamW uses a learning rate of \(5\times10^{-4}\), weight decay \(10^{-3}\), batch size 16, and cosine scheduling. Phases A and C use 30 epochs, while Phase B performs OOF calibration. Attention-MIL follows \citet{ilse2018attentionmil}; complete configurations are provided in the supplementary material.

\begin{table*}[t]
\centering
\caption{
Subject-dependent 10-fold results (mean$\pm$std across test folds).
}
\label{tab:main_10fold}

\begin{tabular}{llcccccc}
\toprule
\textbf{Method}
& \textbf{Metric}
& \textbf{AMIGOS-V}
& \textbf{AMIGOS-A}
& \textbf{MAHNOB-V}
& \textbf{MAHNOB-A}
& \textbf{DEAP-V}
& \textbf{DEAP-A}
\\
\midrule

{Late Fusion}
& Acc
&0.869$\pm$0.042
&0.718$\pm$0.096
&0.790$\pm$0.067
&0.728$\pm$0.070
&0.713$\pm$0.042
&0.631$\pm$0.044
\\
\rowcolor{gray!10}
& F1
&0.880$\pm$0.034
&0.768$\pm$0.084
&0.791$\pm$0.068
&0.711$\pm$0.095
&0.740$\pm$0.043
&0.712$\pm$0.049
\\

{TFN}
& Acc
&0.857$\pm$0.065
&0.731$\pm$0.068
&0.799$\pm$0.056
&0.726$\pm$0.045
&0.684$\pm$0.039
&0.613$\pm$0.061
\\
\rowcolor{gray!10}
& F1
&0.863$\pm$0.065
&0.768$\pm$0.067
&0.801$\pm$0.054
&0.702$\pm$0.058
&0.684$\pm$0.064
&0.676$\pm$0.076
\\

{Attention-MIL}
& Acc
&0.845$\pm$0.065
&0.743$\pm$0.081
&0.801$\pm$0.069
&0.720$\pm$0.082
&0.683$\pm$0.056
&0.627$\pm$0.055
\\
\rowcolor{gray!10}
& F1
&0.855$\pm$0.064
&0.791$\pm$0.083
&0.809$\pm$0.063
&0.711$\pm$0.101
&0.703$\pm$0.056
&0.707$\pm$0.072
\\

{MISA}
& Acc
&0.854$\pm$0.054
&0.745$\pm$0.053
&0.770$\pm$0.102
&0.762$\pm$0.075
&0.695$\pm$0.052
&0.602$\pm$0.039
\\
\rowcolor{gray!10}
& F1
&0.861$\pm$0.057
&0.782$\pm$0.062
&0.760$\pm$0.113
&\textbf{0.760$\pm$0.095}
&0.710$\pm$0.054
&0.692$\pm$0.042
\\

{FuseMoE}
& Acc
&0.866$\pm$0.036
&0.738$\pm$0.067
&0.801$\pm$0.055
&0.742$\pm$0.050
&0.696$\pm$0.051
&0.634$\pm$0.058
\\
\rowcolor{gray!10}
& F1
&0.875$\pm$0.030
&0.788$\pm$0.066
&0.800$\pm$0.061
&0.743$\pm$0.060
&0.718$\pm$0.058
&0.709$\pm$0.068
\\

{EMoE}
& Acc
&0.862$\pm$0.051
&\textbf{0.760$\pm$0.068}
&0.805$\pm$0.052
&0.728$\pm$0.059
&0.682$\pm$0.053
&0.595$\pm$0.060
\\
\rowcolor{gray!10}
& F1
&0.874$\pm$0.041
&0.807$\pm$0.057
&0.800$\pm$0.055
&0.726$\pm$0.067
&0.708$\pm$0.052
&0.681$\pm$0.066
\\

{I2MoE}
& Acc
&0.861$\pm$0.068
&0.748$\pm$0.048
&0.772$\pm$0.067
&0.720$\pm$0.071
&0.685$\pm$0.054
&0.605$\pm$0.068
\\
\rowcolor{gray!10}
& F1
&0.872$\pm$0.061
&0.796$\pm$0.035
&0.782$\pm$0.060
&0.714$\pm$0.095
&0.706$\pm$0.063
&0.691$\pm$0.078
\\

{\textbf{CONFER (Ours)}}
& Acc
&\textbf{0.894$\pm$0.038}
&0.752$\pm$0.068
&\textbf{0.870$\pm$0.050}
&\textbf{0.779$\pm$0.060}
&\textbf{0.722$\pm$0.057}
&\textbf{0.674$\pm$0.030}
\\
\rowcolor{gray!10}
& F1
&\textbf{0.891$\pm$0.037}
&\textbf{0.807$\pm$0.077}
&\textbf{0.810$\pm$0.077}
&0.735$\pm$0.106
&\textbf{0.757$\pm$0.073}
&\textbf{0.714$\pm$0.055}
\\

\bottomrule
\end{tabular}
\end{table*}

\subsection{Main Results}



Under 10-fold CV (Table~\ref{tab:main_10fold}), CONFER achieves the best or competitive results on most tasks, including Accuracy/F1 of 0.894/0.891 on AMIGOS-V and 0.870/0.810 on MAHNOB-V. Under LOSO (Table~\ref{tab:loso_main}), it obtains 0.873/0.879 on AMIGOS-V and 0.854/0.783 on MAHNOB-V. A paired Wilcoxon test over subject-level accuracies confirms a significant improvement over the strongest baseline (\(p<0.05\)).

\begin{table}[t]
\centering
\caption{
Ablation results on AMIGOS under LOSO evaluation. Each entry reports Acc/F1. Same Backbone Fusion shares the CONFER backbone but removes negotiation and regime calibration.
}
\label{tab:ablation_amigos}

\begin{tabular}{lcc}
\toprule
\textbf{Method}
& \textbf{Valence}
& \textbf{Arousal}
\\
\midrule

Same Backbone Fusion
& 0.838/0.846
& 0.600/0.619
\\

Mean Pooling
& 0.844/0.851
& 0.606/0.625
\\

w/o Negotiation
& 0.854/0.860
& 0.616/0.638
\\

w/o Peer Support
& 0.852/0.858
& 0.614/0.633
\\

w/o Sender Reliability
& 0.860/0.866
& 0.621/0.644
\\

w/o Receiver Concession
& 0.863/0.868
& 0.624/0.647
\\

w/o Regime
& 0.849/0.855
& 0.611/0.632
\\

Static Reliability
& 0.842/0.849
& 0.604/0.624
\\

\textbf{Full (CONFER)}
& \textbf{0.873/0.879}
& \textbf{0.632/0.657}
\\

\bottomrule
\end{tabular}
\end{table}

\subsection{Ablation Study}


Table~\ref{tab:ablation_amigos} shows that removing negotiation, sender reliability, or receiver concession consistently degrades LOSO performance, supporting reliability-aware directional interaction. Static Reliability underperforms CONFER, while mean pooling and the removal of peer support or regime calibration cause further reductions.

\begin{figure}[t]
\centering
\includegraphics[width=\linewidth]{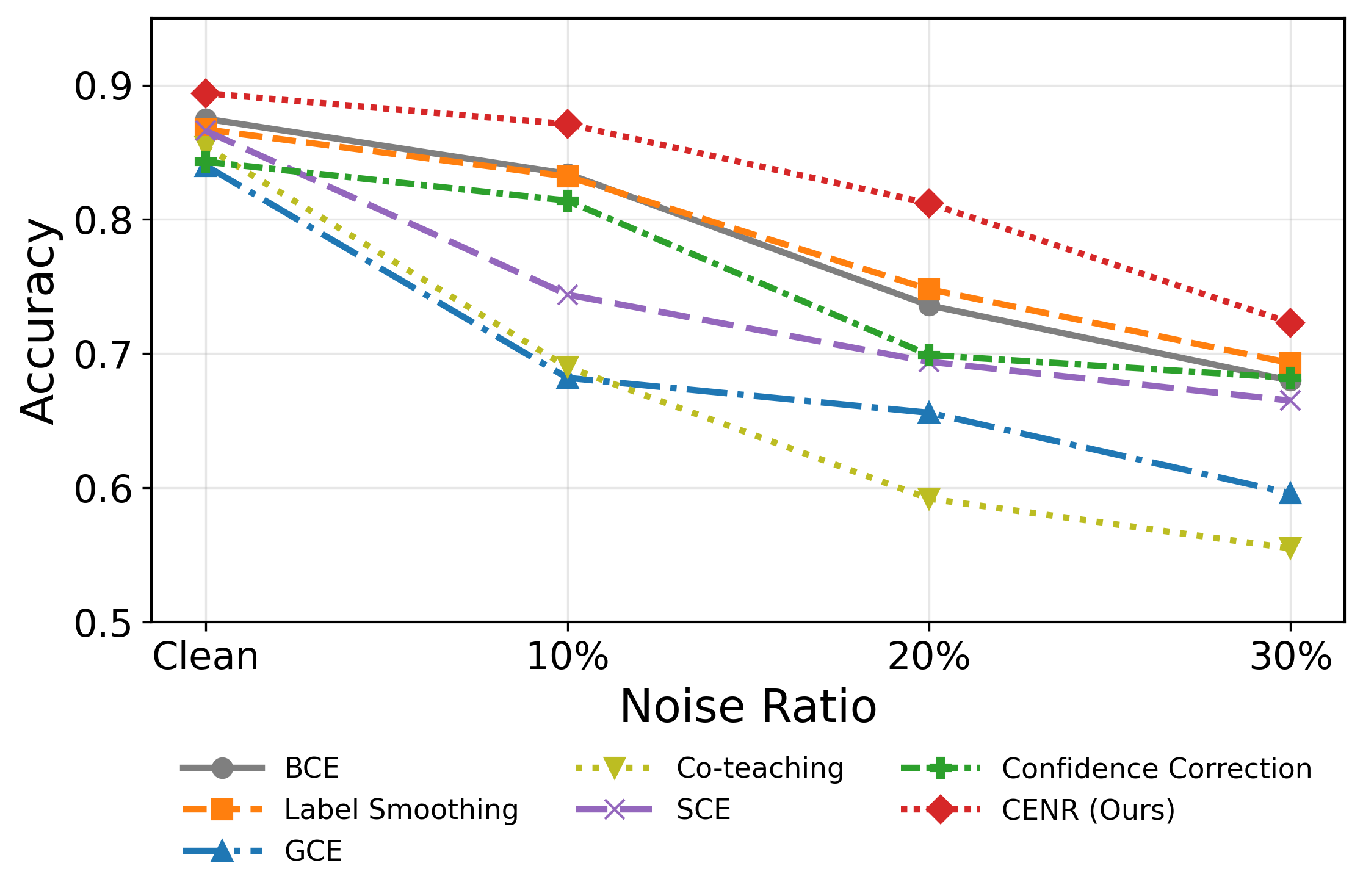}
\caption{
Robustness comparison under symmetric weak-label corruption on AMIGOS-V.
}
\label{fig:noise_robustness}
\end{figure}

\begin{table}[t]
\centering
\small
\setlength{\tabcolsep}{4pt}
\caption{
Robustness under different label corruption settings on AMIGOS-V. 
Entries report Accuracy.
}
\label{tab:structured_noise}

\begin{tabular}{lccc}
\toprule
\textbf{Noise}
& \textbf{Backbone}
& \textbf{w/o Reg.}
& \textbf{CONFER}
\\
\midrule

Clean
& 0.856
& 0.851
& \textbf{0.894}
\\

Sym. 30\%
& 0.650
& 0.658
& \textbf{0.723}
\\

Subj. 30\%
& 0.641
& 0.637
& \textbf{0.686}
\\

\bottomrule
\end{tabular}
\end{table}

\subsection{Robustness to Weak-label Corruption}



As shown in Fig.~\ref{fig:noise_robustness}, all methods degrade as corruption increases, whereas CONFER maintains the most stable trend and achieves 0.723 accuracy under 30\% symmetric corruption. Corrupted samples also exhibit higher Dissent/Ambiguity probability than clean samples (mean soft $\pi_D+\pi_A$: 0.61 vs.\ 0.32), indicating that the inferred regimes capture supervision inconsistency. We further evaluate subject-level corruption by flipping all training labels from 30\% randomly selected subjects while keeping test labels unchanged. Although this setting is more challenging than symmetric noise, CONFER again achieves the best performance and smallest degradation (Table~\ref{tab:structured_noise}).

\begin{table}[t]
\centering
\small
\setlength{\tabcolsep}{3.5pt}
\caption{
Negotiation utility on MAHNOB-HCI (\%). Rescue/Harm denote corrected/disrupted predictions in the high-conflict group.
}
\label{tab:negotiation_utility}

\begin{tabular}{lcccc}
\toprule
\textbf{Target}
& \makecell{\textbf{High-conflict}\\\textbf{$\Delta$Acc}}
& \makecell{\textbf{Low-conflict}\\\textbf{$\Delta$Acc}}
& \textbf{Rescue}
& \textbf{Harm} \\
\midrule

Valence
& +5.4
& +1.0
& 21.9
& 0.37
\\

Arousal
& +8.9
& +0.6
& 38.4
& 2.93
\\

\bottomrule
\end{tabular}
\end{table}

\subsection{Negotiation Utility Analysis}

We rank held-out samples according to their pre-negotiation conflict $D^{(0)}$ and define the upper and lower quartiles as the high- and low-conflict groups, respectively.
As shown in Table~\ref{tab:negotiation_utility}, among high-conflict samples negotiation corrects 21.9\%/38.4\% (valence/arousal) of wrong predictions while disrupting only 0.37\%/2.93\%, i.e.\ rescue exceeds harm by over an order of magnitude. 
Accordingly, negotiation yields accuracy gains of 5.4/8.9 points in the high-conflict quartile, compared with 1.0/0.6 points in the low-conflict quartile, showing that its benefit concentrates on high-conflict samples.

\begin{table}[t]
\centering
\small
\setlength{\tabcolsep}{4pt}
\caption{
Prediction error across inferred conflict regimes on held-out LOSO subjects.
}
\label{tab:regime_error}

\begin{tabular}{lccc}
\toprule
\textbf{Dataset}
& \textbf{Consensus}
& \textbf{Dissent}
& \textbf{Ambiguity}
\\
\midrule

AMIGOS-V
& \textbf{0.071}
& 0.137
& 0.214
\\

MAHNOB-V
& \textbf{0.041}
& 0.161
& 0.200
\\

\bottomrule
\end{tabular}
\end{table}

\subsection{Prediction Reliability Across Conflict Regimes}

Each held-out sample is assigned to the regime with the highest posterior probability among $\pi_C$, $\pi_D$, and $\pi_A$. Table~\ref{tab:regime_error} reports the pooled prediction error within each inferred regime on held-out LOSO subjects. Consensus samples consistently exhibit the lowest error, whereas Dissent and Ambiguity correspond to substantially less reliable predictions. 
This ordering supports the interpretation of the inferred conflict regimes as operational indicators of post-negotiation prediction reliability.
Errors are pooled over held-out samples after seed averaging rather than averaged equally over subjects; therefore, their aggregate values are not expected to equal $1-\mathrm{Acc}$ in Table~\ref{tab:loso_main}.

\begin{table}[t]
\centering
\small
\setlength{\tabcolsep}{6pt}
\caption{
Supervision inconsistency detection AUROC using predictive entropy and negotiation-derived regime factors.
}
\label{tab:regime_confidence}

\begin{tabular}{lcc}
\toprule
\textbf{Features}
& \textbf{AMIGOS-V}
& \textbf{MAHNOB-V}
\\
\midrule

$H$ (Entropy)
& 0.617
& 0.680
\\

$U$
& 0.646
& 0.698
\\

$[R,P,U]$ (LR)
& 0.662
& 0.714
\\

$1-\omega$ (Regime weight)
& 0.653
& 0.711
\\

$\mathbf{[H,R,P,U]}$ (LR)
& \textbf{0.674}
& \textbf{0.718}
\\

\bottomrule
\end{tabular}
\end{table}

\subsection{Supervision-Inconsistency Detection}



We detect supervision-inconsistent held-out samples whose predictions disagree with their weak labels. As shown in Table~\ref{tab:regime_confidence}, predictive entropy $H$ obtains AUROCs of 0.617/0.680, while $1-\omega_i$ and $[R,P,U]$ reach 0.653/0.711 and 0.662/0.714, respectively. Combining all features through a leave-one-fold-out logistic detector achieves the best results of 0.674/0.718, improving over entropy by 0.057/0.038. These results indicate that resolved conflict, residual dissent, and regime weights provide supervision-reliability information beyond predictive confidence.

The frozen Phase-B calibration also remains stable after Phase-C optimization: reliability estimates achieve Pearson/Spearman correlations of 0.976/0.899 with 100\% dominant-modality agreement. The pre-/post-optimization correlation of $\omega_i$ is 0.93, while its correlation with $\tilde{t}_i$ remains 0.82, confirming limited drift in reliability estimation and supervision calibration.

\subsection{Visualization of Negotiation Dynamics}

\begin{figure}[t]
    \centering
    \includegraphics[width=\linewidth]{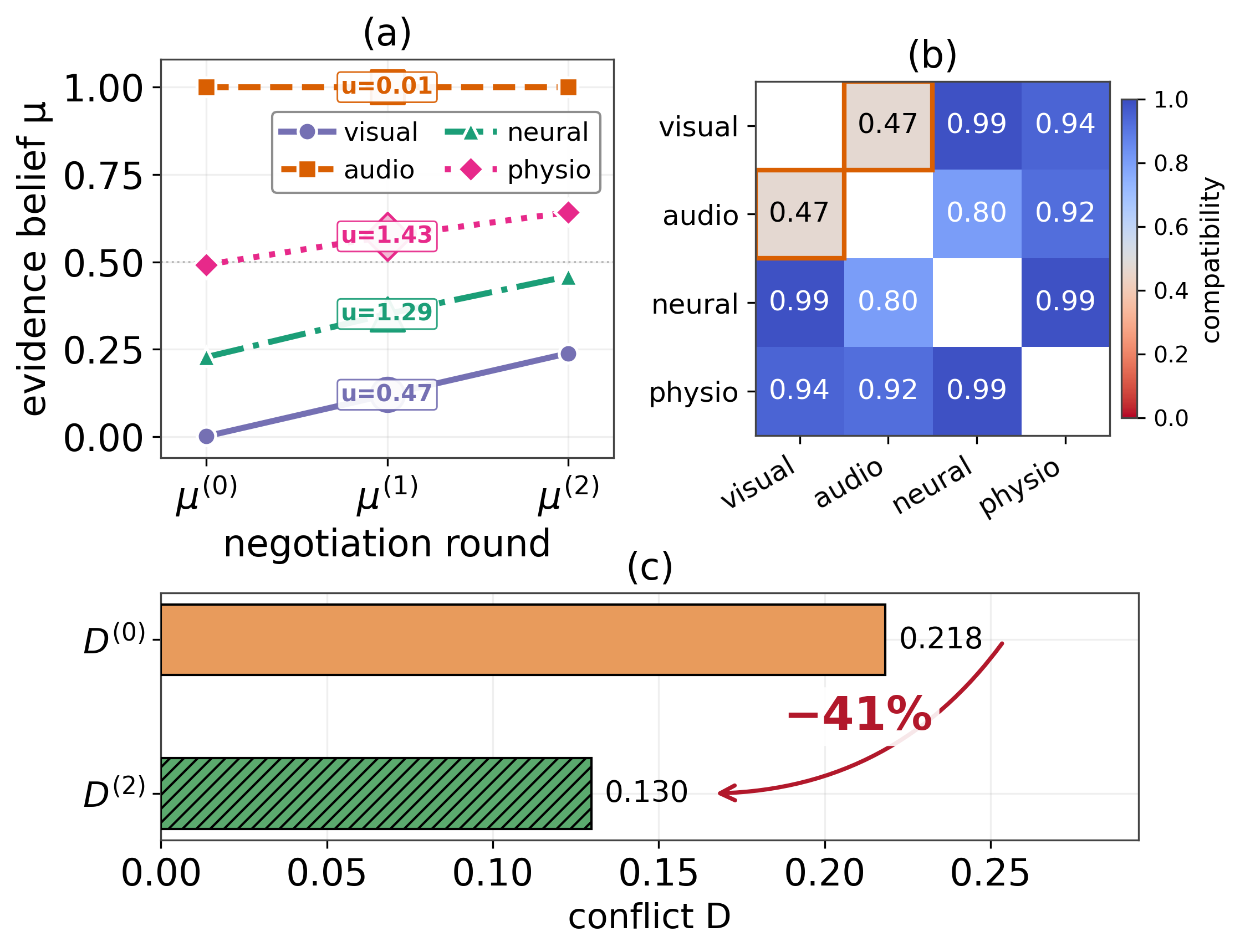}
    \caption{
    CONFER negotiation on a high-conflict sample:
    (a) belief evolution, (b) initial compatibility, and
    (c) conflict reduction.
    }
    \label{fig:case_study}
\end{figure}

Figure~\ref{fig:case_study} visualizes a high-conflict sample in which initially inconsistent modality beliefs are progressively refined through two negotiation rounds. The compatibility matrix and belief trajectories illustrate how CONFER selectively updates experts according to their compatibility and relative reliability.

\section{Discussion}

Unlike direct aggregation or adaptive expert routing~\cite{baltrusaitis2019multimodal,gao2024euar}, CONFER determines both whether a modality provides reliable evidence and whether it should influence another modality for the current sample. Importantly, negotiation does not indiscriminately minimize cross-modal conflict or force experts toward identical predictions. Instead, compatible beliefs from more reliable peers are selectively incorporated, with the larger gains on high-conflict samples indicating that this interaction is most useful when modality evidence is inconsistent. After negotiation, resolved conflict, persistent dissent, and modality uncertainty distinguish the Consensus, Dissent, and Ambiguity regimes. Their different prediction-error patterns and supervision-inconsistency detection performance suggest that these regimes capture information beyond predictive confidence. Together with the corruption results, these findings indicate that cross-modal conflict can serve as an informative signal for both modality coordination and weak-label reliability estimation.

\section{Conclusion}



This paper introduced CONFER, a graph-based conflict-aware framework for weakly supervised multimodal emotion recognition. Reliability-aware directional message passing refines modality predictions, while negotiation-derived conflict regimes calibrate each sample's contribution to weak-label training. Experiments on AMIGOS, MAHNOB-HCI, and DEAP demonstrate competitive performance across evaluation protocols and improved robustness to weak-label corruption. These results show that cross-modal conflict can serve not only as a source of prediction inconsistency, but also as an informative signal for modality coordination and supervision-reliability assessment.

\bibliography{aaai2027}

@article{russell1980circumplex,
  author  = {Russell, James A.},
  title   = {A Circumplex Model of Affect},
  journal = {Journal of Personality and Social Psychology},
  year    = {1980},
  volume  = {39},
  number  = {6},
  pages   = {1161--1178},
  doi     = {10.1037/h0077714},
  url     = {https://doi.org/10.1037/h0077714}
}

@article{bradley1994sam,
  author  = {Bradley, Margaret M. and Lang, Peter J.},
  title   = {Measuring Emotion: The Self-Assessment Manikin and the Semantic Differential},
  journal = {Journal of Behavior Therapy and Experimental Psychiatry},
  year    = {1994},
  volume  = {25},
  number  = {1},
  pages   = {49--59},
  doi     = {10.1016/0005-7916(94)90063-9},
  url     = {https://doi.org/10.1016/0005-7916(94)90063-9}
}

@article{mauss2009measures,
  author  = {Mauss, Iris B. and Robinson, Michael D.},
  title   = {Measures of Emotion: A Review},
  journal = {Cognition and Emotion},
  year    = {2009},
  volume  = {23},
  number  = {2},
  pages   = {209--237},
  doi     = {10.1080/02699930802204677},
  url     = {https://doi.org/10.1080/02699930802204677}
}

@article{barrett2017constructed,
  author  = {Barrett, Lisa Feldman},
  title   = {The Theory of Constructed Emotion: An Active Inference Account of Interoception and Categorization},
  journal = {Social Cognitive and Affective Neuroscience},
  year    = {2017},
  volume  = {12},
  number  = {1},
  pages   = {1--23},
  doi     = {10.1093/scan/nsw154},
  url     = {https://doi.org/10.1093/scan/nsw154}
}

@article{koelstra2012deap,
  author  = {Koelstra, Sander and Muehl, Christian and Soleymani, Mohammad and Lee, Jong-Seok and Yazdani, Ashkan and Ebrahimi, Touradj and Pun, Thierry and Nijholt, Anton and Patras, Ioannis},
  title   = {{DEAP}: A Database for Emotion Analysis Using Physiological Signals},
  journal = {IEEE Transactions on Affective Computing},
  year    = {2012},
  volume  = {3},
  number  = {1},
  pages   = {18--31},
  doi     = {10.1109/T-AFFC.2011.15},
  url     = {https://doi.org/10.1109/T-AFFC.2011.15}
}

@article{soleymani2012mahnob,
  author  = {Soleymani, Mohammad and Lichtenauer, Jeroen and Pun, Thierry and Pantic, Maja},
  title   = {A Multimodal Database for Affect Recognition and Implicit Tagging},
  journal = {IEEE Transactions on Affective Computing},
  year    = {2012},
  volume  = {3},
  number  = {1},
  pages   = {42--55},
  doi     = {10.1109/T-AFFC.2011.25},
  url     = {https://doi.org/10.1109/T-AFFC.2011.25}
}

@article{miranda2021amigos,
  author  = {Miranda-Correa, Juan Abdon and Abadi, Mojtaba Khomami and Sebe, Nicu and Patras, Ioannis},
  title   = {{AMIGOS}: A Dataset for Affect, Personality and Mood Research on Individuals and Groups},
  journal = {IEEE Transactions on Affective Computing},
  year    = {2021},
  volume  = {12},
  number  = {2},
  pages   = {479--493},
  doi     = {10.1109/TAFFC.2018.2884461},
  url     = {https://doi.org/10.1109/TAFFC.2018.2884461}
}

@article{baltrusaitis2019multimodal,
  author  = {Baltru\v{s}aitis, Tadas and Ahuja, Chaitanya and Morency, Louis-Philippe},
  title   = {Multimodal Machine Learning: A Survey and Taxonomy},
  journal = {IEEE Transactions on Pattern Analysis and Machine Intelligence},
  year    = {2019},
  volume  = {41},
  number  = {2},
  pages   = {423--443},
  doi     = {10.1109/TPAMI.2018.2798607},
  url     = {https://doi.org/10.1109/TPAMI.2018.2798607}
}

@inproceedings{zadeh2017tfn,
  author    = {Zadeh, Amir and Chen, Minghai and Poria, Soujanya and Cambria, Erik and Morency, Louis-Philippe},
  title     = {Tensor Fusion Network for Multimodal Sentiment Analysis},
  booktitle = {Proceedings of the 2017 Conference on Empirical Methods in Natural Language Processing},
  year      = {2017},
  pages     = {1103--1114},
  doi       = {10.18653/v1/D17-1115},
  url       = {https://aclanthology.org/D17-1115/}
}

@inproceedings{tsai2019mult,
  author    = {Tsai, Yao-Hung Hubert and Bai, Shaojie and Liang, Paul Pu and Kolter, J. Zico and Morency, Louis-Philippe and Salakhutdinov, Ruslan},
  title     = {Multimodal Transformer for Unaligned Multimodal Language Sequences},
  booktitle = {Proceedings of the 57th Annual Meeting of the Association for Computational Linguistics},
  year      = {2019},
  pages     = {6558--6569},
  doi       = {10.18653/v1/P19-1656},
  url       = {https://aclanthology.org/P19-1656/}
}

@inproceedings{hazarika2020misa,
  author    = {Hazarika, Devamanyu and Zimmermann, Roger and Poria, Soujanya},
  title     = {{MISA}: Modality-Invariant and -Specific Representations for Multimodal Sentiment Analysis},
  booktitle = {Proceedings of the 28th ACM International Conference on Multimedia},
  year      = {2020},
  pages     = {1122--1131},
  doi       = {10.1145/3394171.3413678},
  url       = {https://doi.org/10.1145/3394171.3413678}
}

@inproceedings{yu2021selfmm,
  author    = {Yu, Wenmeng and Xu, Hua and Yuan, Ziqi and Wu, Jiele},
  title     = {Learning Modality-Specific Representations with Self-Supervised Multi-Task Learning for Multimodal Sentiment Analysis},
  booktitle = {Proceedings of the AAAI Conference on Artificial Intelligence},
  year      = {2021},
  volume    = {35},
  pages     = {10790--10797},
  doi       = {10.1609/aaai.v35i12.17289},
  url       = {https://doi.org/10.1609/aaai.v35i12.17289}
}

@inproceedings{han2021mmim,
  author    = {Han, Wei and Chen, Hui and Poria, Soujanya},
  title     = {Improving Multimodal Fusion with Hierarchical Mutual Information Maximization for Multimodal Sentiment Analysis},
  booktitle = {Proceedings of the 2021 Conference on Empirical Methods in Natural Language Processing},
  year      = {2021},
  pages     = {9180--9192},
  doi       = {10.18653/v1/2021.emnlp-main.723},
  url       = {https://aclanthology.org/2021.emnlp-main.723/}
}

@inproceedings{mittal2020m3er,
  author    = {Mittal, Trisha and Bhattacharya, Uttaran and Chandra, Rohan and Bera, Aniket and Manocha, Dinesh},
  title     = {{M3ER}: Multiplicative Multimodal Emotion Recognition Using Facial, Textual, and Speech Cues},
  booktitle = {Proceedings of the AAAI Conference on Artificial Intelligence},
  year      = {2020},
  volume    = {34},
  pages     = {1359--1367},
  doi       = {10.1609/aaai.v34i02.5492},
  url       = {https://doi.org/10.1609/aaai.v34i02.5492}
}

@inproceedings{rayatdoost2020gated,
  author    = {Rayatdoost, Soheil and Rudrauf, David and Soleymani, Mohammad},
  title     = {Multimodal Gated Information Fusion for Emotion Recognition from {EEG} Signals and Facial Behaviors},
  booktitle = {Proceedings of the 2020 International Conference on Multimodal Interaction},
  year      = {2020},
  pages     = {655--659},
  doi       = {10.1145/3382507.3418864},
  url       = {https://doi.org/10.1145/3382507.3418864}
}

@article{zhang2021regularized,
  title={Emotion recognition from multimodal physiological signals using a regularized deep fusion of kernel machine},
  author={Zhang, Xiaowei and Liu, Jinyong and Shen, Jian and Li, Shaojie and Hou, Kechen and Hu, Bin and Gao, Jin and Zhang, Tong},
  journal={IEEE transactions on cybernetics},
  volume={51},
  number={9},
  pages={4386--4399},
  year={2020},
  publisher={IEEE}
}

@article{liu2024missing,
  author  = {Liu, Rui and Zuo, Haolin and Lian, Zheng and Schuller, Bj\"orn W. and Li, Haizhou},
  title   = {Contrastive Learning Based Modality-Invariant Feature Acquisition for Robust Multimodal Emotion Recognition With Missing Modalities},
  journal = {IEEE Transactions on Affective Computing},
  year    = {2024},
  volume  = {15},
  pages   = {1856--1873},
  url     = {https://dblp.org/rec/journals/taffco/LiuZLSL24}
}

@inproceedings{han2024fusemoe,
  author    = {Han, Xing and Nguyen, Huy and Harris, Carl and Ho, Nhat and Saria, Suchi},
  title     = {{FuseMoE}: Mixture-of-Experts Transformers for Fleximodal Fusion},
  booktitle = {Advances in Neural Information Processing Systems},
  year      = {2024},
  volume    = {37},
  url       = {https://proceedings.neurips.cc/paper_files/paper/2024/hash/7d62a85ebfed2f680eb5544beae93191-Abstract-Conference.html}
}

@inproceedings{fang2025emoe,
  author    = {Fang, Yiyang and Huang, Wenke and Wan, Guancheng and Su, Kehua and Ye, Mang},
  title     = {{EMOE}: Modality-Specific Enhanced Dynamic Emotion Experts},
  booktitle = {Proceedings of the IEEE/CVF Conference on Computer Vision and Pattern Recognition},
  year      = {2025},
  pages     = {14314--14324},
  url       = {https://openaccess.thecvf.com/content/CVPR2025/html/Fang_EMOE_Modality-Specific_Enhanced_Dynamic_Emotion_Experts_CVPR_2025_paper.html}
}

@inproceedings{xin2025i2moe,
  author    = {Xin, Jiayi and Yun, Sukwon and Peng, Jie and Choi, Inyoung and Ballard, Jenna L. and Chen, Tianlong and Long, Qi},
  title     = {{I2MoE}: Interpretable Multimodal Interaction-Aware Mixture-of-Experts},
  booktitle = {Proceedings of the 42nd International Conference on Machine Learning},
  year      = {2025},
  url       = {https://openreview.net/forum?id=EuJaF5QsMP}
}

@inproceedings{gao2024euar,
  author    = {Gao, Zixian and Hu, Disen and Jiang, Xun and Lu, Huimin and Shen, Heng Tao and Xu, Xing},
  title     = {Enhanced Experts with Uncertainty-Aware Routing for Multimodal Sentiment Analysis},
  booktitle = {Proceedings of the 32nd ACM International Conference on Multimedia},
  year      = {2024},
  pages     = {9650--9659},
  doi       = {10.1145/3664647.3680949},
  url       = {https://doi.org/10.1145/3664647.3680949}
}

@inproceedings{zeng2022mitigating,
  title={Mitigating inconsistencies in multimodal sentiment analysis under uncertain missing modalities},
  author={Zeng, Jiandian and Zhou, Jiantao and Liu, Tianyi},
  booktitle={Proceedings of the 2022 conference on empirical methods in natural language processing},
  pages={2924--2934},
  year={2022}
}

@inproceedings{guo2017calibration,
  author    = {Guo, Chuan and Pleiss, Geoff and Sun, Yu and Weinberger, Kilian Q.},
  title     = {On Calibration of Modern Neural Networks},
  booktitle = {Proceedings of the 34th International Conference on Machine Learning},
  year      = {2017},
  pages     = {1321--1330},
  url       = {https://proceedings.mlr.press/v70/guo17a.html}
}

@inproceedings{kendall2017uncertainties,
  author    = {Kendall, Alex and Gal, Yarin},
  title     = {What Uncertainties Do We Need in Bayesian Deep Learning for Computer Vision?},
  booktitle = {Advances in Neural Information Processing Systems},
  year      = {2017},
  volume    = {30},
  url       = {https://proceedings.neurips.cc/paper/2017/hash/2650d6089a6d640c5e85b2b88265dc2b-Abstract.html}
}

@inproceedings{sensoy2018evidential,
  author    = {Sensoy, Murat and Kaplan, Lance and Kandemir, Melih},
  title     = {Evidential Deep Learning to Quantify Classification Uncertainty},
  booktitle = {Advances in Neural Information Processing Systems},
  year      = {2018},
  volume    = {31},
  url       = {https://proceedings.neurips.cc/paper/2018/hash/a981f2b708044d6fb4a71a1463242520-Abstract.html}
}

@inproceedings{han2021tmc,
  author    = {Han, Zongbo and Zhang, Changqing and Fu, Huazhu and Zhou, Joey Tianyi},
  title     = {Trusted Multi-View Classification},
  booktitle = {International Conference on Learning Representations},
  year      = {2021},
  url       = {https://openreview.net/forum?id=OOsR8BzCnl5}
}

@article{han2023dynamic,
  author  = {Han, Zongbo and Zhang, Changqing and Fu, Huazhu and Zhou, Joey Tianyi},
  title   = {Trusted Multi-View Classification With Dynamic Evidential Fusion},
  journal = {IEEE Transactions on Pattern Analysis and Machine Intelligence},
  year    = {2023},
  volume  = {45},
  number  = {2},
  pages   = {2551--2566},
  doi     = {10.1109/TPAMI.2022.3171983},
  url     = {https://doi.org/10.1109/TPAMI.2022.3171983}
}

@article{song2023noisylabels,
  author  = {Song, Hwanjun and Kim, Minseok and Park, Dongmin and Shin, Yooju and Lee, Jae-Gil},
  title   = {Learning From Noisy Labels With Deep Neural Networks: A Survey},
  journal = {IEEE Transactions on Neural Networks and Learning Systems},
  year    = {2023},
  volume  = {34},
  number  = {11},
  pages   = {8135--8153},
  doi     = {10.1109/TNNLS.2022.3152527},
  url     = {https://doi.org/10.1109/TNNLS.2022.3152527}
}

@inproceedings{zhang2018gce,
  author    = {Zhang, Zhilu and Sabuncu, Mert R.},
  title     = {Generalized Cross Entropy Loss for Training Deep Neural Networks with Noisy Labels},
  booktitle = {Advances in Neural Information Processing Systems},
  year      = {2018},
  volume    = {31},
  url       = {https://proceedings.neurips.cc/paper/2018/hash/f2925f97bc13ad2852a7a551802feea0-Abstract.html}
}

@inproceedings{wang2019sce,
  author    = {Wang, Yisen and Ma, Xingjun and Chen, Zaiyi and Luo, Yuan and Yi, Jinfeng and Bailey, James},
  title     = {Symmetric Cross Entropy for Robust Learning With Noisy Labels},
  booktitle = {Proceedings of the IEEE/CVF International Conference on Computer Vision},
  year      = {2019},
  pages     = {322--330},
  url       = {https://openaccess.thecvf.com/content_ICCV_2019/html/Wang_Symmetric_Cross_Entropy_for_Robust_Learning_With_Noisy_Labels_ICCV_2019_paper.html}
}

@inproceedings{han2018coteaching,
  author    = {Han, Bo and Yao, Quanming and Yu, Xingrui and Niu, Gang and Xu, Miao and Hu, Weihua and Tsang, Ivor and Sugiyama, Masashi},
  title     = {Co-Teaching: Robust Training of Deep Neural Networks with Extremely Noisy Labels},
  booktitle = {Advances in Neural Information Processing Systems},
  year      = {2018},
  volume    = {31},
  url       = {https://proceedings.neurips.cc/paper/2018/hash/a19744e268754fb0148b017647355b7b-Abstract.html}
}

@inproceedings{li2020dividemix,
  author    = {Li, Junnan and Socher, Richard and Hoi, Steven C. H.},
  title     = {{DivideMix}: Learning with Noisy Labels as Semi-Supervised Learning},
  booktitle = {International Conference on Learning Representations},
  year      = {2020},
  url       = {https://openreview.net/forum?id=HJgExaVtwr}
}

@inproceedings{liu2020elr,
  author    = {Liu, Sheng and Niles-Weed, Jonathan and Razavian, Narges and Fernandez-Granda, Carlos},
  title     = {Early-Learning Regularization Prevents Memorization of Noisy Labels},
  booktitle = {Advances in Neural Information Processing Systems},
  year      = {2020},
  volume    = {33},
  pages     = {20331--20342},
  url       = {https://proceedings.neurips.cc/paper/2020/hash/ea89621bee7c88b2c5be6681c8ef4906-Abstract.html}
}

@inproceedings{xia2022uncertainty,
  author    = {Xia, Xiaobo and Liu, Tongliang and Han, Bo and Gong, Mingming and Yu, Jun and Niu, Gang and Sugiyama, Masashi},
  title     = {Sample Selection with Uncertainty of Losses for Learning with Noisy Labels},
  booktitle = {International Conference on Learning Representations},
  year      = {2022},
  url       = {https://openreview.net/forum?id=xENf4QUL4LW}
}

@inproceedings{li2022selcl,
  author    = {Li, Shikun and Xia, Xiaobo and Ge, Shiming and Liu, Tongliang},
  title     = {Selective-Supervised Contrastive Learning With Noisy Labels},
  booktitle = {Proceedings of the IEEE/CVF Conference on Computer Vision and Pattern Recognition},
  year      = {2022},
  pages     = {316--325},
  url       = {https://openaccess.thecvf.com/content/CVPR2022/html/Li_Selective-Supervised_Contrastive_Learning_With_Noisy_Labels_CVPR_2022_paper.html}
}

@article{shu2023cmwnet,
  title={Cmw-net: Learning a class-aware sample weighting mapping for robust deep learning},
  author={Shu, Jun and Yuan, Xiang and Meng, Deyu and Xu, Zongben},
  journal={IEEE Transactions on Pattern Analysis and Machine Intelligence},
  volume={45},
  number={10},
  pages={11521--11539},
  year={2023},
  publisher={IEEE}
}

@inproceedings{huang2023twin,
  author    = {Huang, Zhizhong and Zhang, Junping and Shan, Hongming},
  title     = {Twin Contrastive Learning With Noisy Labels},
  booktitle = {Proceedings of the IEEE/CVF Conference on Computer Vision and Pattern Recognition},
  year      = {2023},
  pages     = {11661--11670},
  url       = {https://openaccess.thecvf.com/content/CVPR2023/html/Huang_Twin_Contrastive_Learning_With_Noisy_Labels_CVPR_2023_paper.html}
}

@inproceedings{wang2024noisegpt,
  author    = {Wang, Haoyu and Huang, Zhuo and Lin, Zhiwei and Liu, Tongliang},
  title     = {{NoiseGPT}: Label Noise Detection and Rectification through Probability Curvature},
  booktitle = {Advances in Neural Information Processing Systems},
  year      = {2024},
  volume    = {37},
  url       = {https://proceedings.neurips.cc/paper_files/paper/2024/hash/d95cb79a3421e6d9b6c9a9008c4d07c5-Abstract-Conference.html}
}

@inproceedings{kim2024confrag,
  author    = {Kim, Chris Dongjoo and Moon, Sangwoo and Moon, Jihwan and Woo, Dongyeon and Kim, Gunhee},
  title     = {Sample Selection via Contrastive Fragmentation for Noisy Label Regression},
  booktitle = {Advances in Neural Information Processing Systems},
  year      = {2024},
  volume    = {37},
  url       = {https://proceedings.neurips.cc/paper_files/paper/2024/hash/e68c3a624d19154b28951e8690834607-Abstract-Conference.html}
}

@inproceedings{ilse2018attentionmil,
  author    = {Ilse, Maximilian and Tomczak, Jakub and Welling, Max},
  title     = {Attention-Based Deep Multiple Instance Learning},
  booktitle = {Proceedings of the 35th International Conference on Machine Learning},
  year      = {2018},
  pages     = {2127--2136},
  url       = {https://proceedings.mlr.press/v80/ilse18a.html}
}

@article{kapoor2023leakage,
  author  = {Kapoor, Sayash and Narayanan, Arvind},
  title   = {Leakage and the Reproducibility Crisis in Machine-Learning-Based Science},
  journal = {Patterns},
  year    = {2023},
  volume  = {4},
  number  = {9},
  pages   = {100804},
  doi     = {10.1016/j.patter.2023.100804},
  url     = {https://doi.org/10.1016/j.patter.2023.100804}
}

@inproceedings{bezirganyan2025discounted,
  title     = {Multimodal Learning with Uncertainty Quantification Based on Discounted Belief Fusion},
  author    = {Bezirganyan, Grigor and Sellami, Sana and Berti-Equille, Laure and Fournier, S{\'e}bastien},
  booktitle = {Proceedings of the 28th International Conference on Artificial Intelligence and Statistics},
  series    = {Proceedings of Machine Learning Research},
  volume    = {258},
  pages     = {3142--3150},
  year      = {2025},
  publisher = {PMLR},
  url       = {https://proceedings.mlr.press/v258/bezirganyan25a.html}
}

@inproceedings{lu2025navigating,
  title     = {Navigating Conflicting Views: Harnessing Trust for Learning},
  author    = {Lu, Jueqing and Buntine, Wray and Qi, Yuanyuan and Dipnall, Joanna and Gabbe, Belinda and Du, Lan},
  booktitle = {Proceedings of the 42nd International Conference on Machine Learning},
  series    = {Proceedings of Machine Learning Research},
  volume    = {267},
  pages     = {40411--40435},
  year      = {2025},
  publisher = {PMLR},
  url       = {https://proceedings.mlr.press/v267/lu25a.html}
}


\end{document}